%% file: main.tex
\documentclass[letterpaper]{article} %
\usepackage[preprint]{aaai2027}  %
\usepackage[hyphens]{url}  %
\usepackage{graphicx} %
\usepackage{natbib}  %
\usepackage{caption} %
\usepackage{algorithm}
\usepackage{algorithmic}
\usepackage{multirow}

\usepackage{newfloat}
\usepackage{listings}
\DeclareCaptionStyle{ruled}{labelfont=normalfont,labelsep=colon,strut=off} %
\floatstyle{ruled}
\newfloat{listing}{tb}{lst}{}
\floatname{listing}{Listing}

\usepackage{booktabs}

\usepackage{subcaption}
\usepackage{xcolor}
\usepackage{amsmath}
\usepackage{amssymb}
\usepackage{microtype}
\usepackage{tabularx}
\usepackage[inline]{enumitem}

\definecolor{myred}{HTML}{E31A1C}

\title{Foundation Models are Implicit Deepfake Detectors}
\author{
    Written by AAAI Press Staff\textsuperscript{\rm 1}\thanks{With help from the AAAI Publications Committee.}\\
    AAAI Style Contributions by Peter Patel Schneider,
    Sunil Issar,\\
    J. Scott Penberthy,
    George Ferguson,
    Hans Guesgen,
    Francisco Cruz\equalcontrib\corresponding,
    Marc Pujol-Gonzalez\equalcontrib\corresponding
}
\affiliations{
    \textsuperscript{\rm 1}Association for the Advancement of Artificial Intelligence\\

    1101 Pennsylvania Ave, NW Suite 300\\
    Washington, DC 20004 USA\\
    proceedings-questions@aaai.org
}

\title{Foundation Models are Implicit Deepfake Detectors}
\author {
    Stefan Smeu\textsuperscript{\rm 1},
    Dragos-Alexandru Boldisor\textsuperscript{\rm 1,\rm 2},
    Elisabeta Oneata\textsuperscript{\rm 1},
    Dan Oneata\textsuperscript{\rm 1,\rm 2}
}
\affiliations {
    \textsuperscript{\rm 1}Bitdefender, Romania \\
    \textsuperscript{\rm 2}POLITEHNICA Bucharest \\
}

\begin{document}

\maketitle

\begin{abstract}
Pretrained self-supervised representations have emerged as a core component of current deepfake detection methods, yet it remains unclear which of their properties make real and fake media distinguishable.
In this work, we uncover a surprisingly consistent phenomenon: across multiple pretrained models, datasets, and both image and video domains, fake samples systematically produce lower-magnitude representations than their real counterparts. 
Motivated by this finding, we formulate deepfake detection as an anomaly detection problem and show that simple statistics of feature magnitude achieve competitive performance with far more sophisticated deepfake detection methods.
We further investigate the origin of this effect and demonstrate that reduced feature magnitude is primarily associated with semantic shifts introduced by fake content, while low-level generative fingerprints play a comparatively smaller role. 
Finally, we show that this discriminative signal strengthens as the size of the underlying foundation model grows, suggesting that advances in representation learning naturally translate into stronger zero-shot deepfake detectors.

\end{abstract}

\section{Introduction}

Deepfake detection aims to automatically distinguish synthetic content (produced by generative models) from authentic media content (image, video, audio).
A wide range of approaches are continuously being proposed for this task.
These range from classifiers that use low-level forensic artifacts \cite{marra2019gans,frank2020leveraging,chai2020eccv,tan2024frequency} to methods that rely on semantic inconsistencies,
such as physiological cues---blinking \cite{li2018ictu}, pulse signals \cite{ciftci2020fakecatcher}, breathing \citealt{layton2025every})---or
physical cues---lighting \cite{carvalho2015exposing}, head pose \cite{yang2019exposing}, scene geometry \cite{sarkar2024shadows}.

However, modern detectors tend to no longer engineer specific semantic cues;
instead, they rely on pretrained self-supervised representations,
such as CLIP \cite{radford2021learning}, AV-HuBERT \cite{shi2022learning}, wav2vec 2.0 \cite{baevski2020wav2vec}, depending on the modalities involved.
These supervised representations encode rich high-level information that has proved effective for discriminating fake from real samples.
The simplest approach uses frozen pretrained features with a downstream classifier \cite{ojha2023cvpr,koutlis2024leveraging}.
Other methods further improve performance by fine-tuning or adapting the pretrained representations \cite{khan2024clipping,park2025community},
or by learning representations on data and pretext tasks that are more closely aligned with deepfake detection \cite{haliassos2022leveraging,feng2023self,oorloff2024avff}.
Regardless of the specific strategy, these approaches almost invariably rely on an additional classifier---often a complex one---to separate reals from fakes.

\begin{figure}[t]
\centering
\begin{subfigure}{0.49\columnwidth}
    \includegraphics[width=\linewidth]{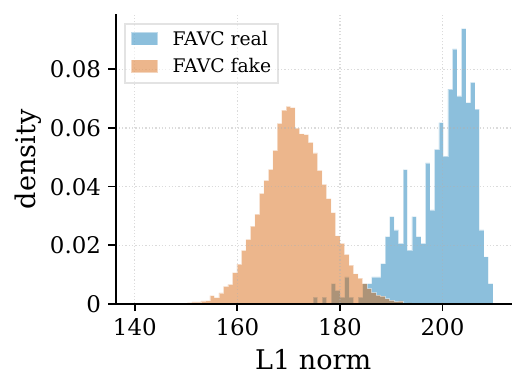} 
    \caption{Videos -- visual AV-HuBERT}
\end{subfigure}
\hfill
\begin{subfigure}{0.49\columnwidth}
    \includegraphics[width=\linewidth]{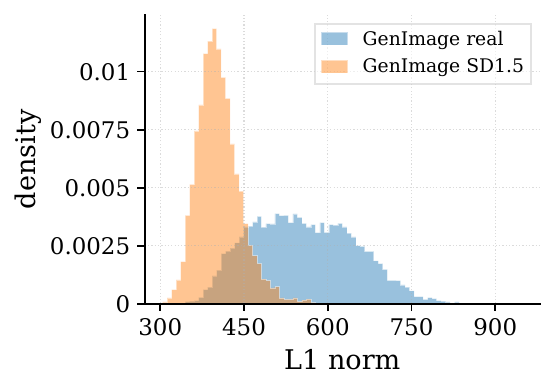} 
    \caption{Images -- DINOv3-7B}
\end{subfigure}

\caption{Histograms of $\ell_1$ feature norm. They are shown for real (blue) and fake (orange) samples extracted from (a) visual-only AV-HuBERT on FakeAVCeleb and (b) DINOv3-7B on GenImage SD1.5. In both settings, fake samples exhibit features with lower $\ell_1$ norm.}
\label{fig:main_l1_figure}
\end{figure}

In this paper, we ask a much simpler question:
can self-supervised features be used \textit{directly} for deepfake detection,
without learning a classifier at all?
Surprisingly, we find that the answer is yes.
Concretely, we show that the $\ell_1$ norm of self-supervised features alone is sufficient to distinguish real from fake samples to a large degree.
We find that this observation holds across two domains (image and video) and across different self-supervised models such as DINOv3 \cite{simeoni2025dinov3} and AV-HuBERT \cite{shi2022learning}, as illustrated in Figure~\ref{fig:main_l1_figure}.

Although this finding may initially seem surprising, it is closely related to observations from the out-of-distribution detection literature.
Previous work has shown that classifiers tend to produce feature representations with larger norms for in-distribution samples and smaller norms for out-of-distribution inputs \cite{dhamija2018reducing,dietterich2022familiarity,Park_2023_ICCV}.
While self-supervised models are not trained as classifiers, we observe an analogous phenomenon.
Because they are pretrained on massive collections of real data, they learn representations that encode real samples more confidently, resulting in larger feature norms.
In contrast, fake samples lie outside the training distribution, producing weaker representations with smaller norms.
This naturally enables feature norms to serve as an effective signal for deepfake detection.

Furthermore, we show that this principle can be refined by using the ratio between the $\ell_1$ and $\ell_2$ norms of the feature representation. 
This encodes the sparsity of the features~\cite{hoyer2004nmf,lopes2013estimating} and represents an even more discriminative score, outperforming existing state-of-the-art methods across a wide variety of deepfake detection datasets.
We also find that the best results are obtained with features extracted from the later layers of the models, and that the size of the self-supervised model correlates with downstream performance.
Overall, our results show that self-supervised features possess a previously underappreciated anomaly detection capability, showing that they can be used directly for deepfake detection without requiring an explicit learned classifier.

To summarize, our contributions are as follows:
\begin{enumerate*}[label=(\roman*)]
    \item We uncover a previously overlooked property of pretrained visual representations:
    fake content produces lower-magnitude features than real content,
    across diverse self-supervised models, datasets, and both image and video domains.
    \item We investigate the origin of this phenomenon and demonstrate that 
    it
    is primarily driven by the semantic shift introduced by fake content
    rather than low-level fingerprints.
    \item
    We show that simple statistics based on feature magnitude enable competitive performance,
    making this a strong baseline for deepfake detection.
\end{enumerate*}

\section{Related Work}

While the de facto approach to deepfake detection frames the task as binary classification,
this often suffers from poor generalization to unseen manipulations \cite{ricker2024towards} and a reliance on shortcut features \cite{AVH-Align}.
As such, a promising alternative is to assume access only to real samples during training.
We identify three main directions of using real samples only to build deepfake detectors.

\paragraph{Anomaly detection.}
The most direct approach models the distribution of real samples and considers as fake any sample that deviate from the learned real manifold.
This idea has been particularly successful in the video domain, where inconsistencies between the audio and visual streams can reveal manipulated content.
Such misalignments can be measured directly using an audio-visual foundation model by computing the cosine similarity between audio and visual features \cite{liang24speechforensics,reiss2023detectingdeepfakesseeing}.
Alternatively, a dedicated model can be trained on top of these features using only real data \cite{AVH-Align}.
Moreover, \citet{feng2023self} observe that real video also exhibits audio-visual de-synchronizations, and therefore explicitly model the distribution of de-synchronizations of real video rather than treating every such misalignment as evidence of manipulation.

\paragraph{Reconstruction-based approaches.}
A different line of work is to use generative models to model the \textit{fake} manifold.
Even if generative models are trained on real data only, since they were used to generate fakes, they can model this distribution.
The main intuition is that fake samples are typically better reconstructed than real ones.
Concretely, a sample is scored by the $\ell_2$ distance between a sample and its reconstruction through the generative model \cite{ricker_2024_CVPR}.
This idea has been further refined, for example, by replacing the reconstruction error with a curvature-based metric in the latent space \cite{brokman2025manifold}, or by evaluating reconstruction consistency across perturbed versions of the input (e.g., rotations or blurring) \cite{choi2026debiased}.
A key limitation of these approaches is that their effectiveness depends on the overlap between the generators used to compute the reconstructions and those used to generate fake samples.

\paragraph{Consistency-based approaches.}
A more recent line of work is based on the observation that manipulation artifacts are often unstable under input perturbations.
These methods apply a transformation to an image and measure the similarity between the original and perturbed samples in a learned feature space.
Since real images are generally more stable, they tend to produce higher similarities than fake images \cite{he2024rigid, tsai2024minder, choi2025warpad}.
These approaches rely on the DINOv2 feature extractor \cite{oquab2023dinov2} and compute the cosine similarity between an image and its perturbation.
Perturbations include 
Gaussian noise \cite{he2024rigid},
blurring and sharpening \cite{tsai2024minder},
high-frequency transformations \cite{choi2025warpad}.

\par\vspace{6pt}\noindent%
Our method is much simpler and direct than those presented:
it produces a fakeness score based on the feature norm of self-supervised representations.
As such, it shares characteristics with anomaly detection methods.
Closest to our approach is STALL \cite{ben2026stall}, which computes the distance of features in the whitening space to the real distribution, using the $\ell_2$ norm.
However, this method requires an additional calibration stage using a held-out real dataset.

\section{Methodology}

\paragraph{Magnitude differs between real and fake samples.}
We start with the observation that pretrained foundation models consistently produce lower-magnitude representations for fake samples than for authentic ones. %
To show this, we extract visual representations from the AV-HuBERT \cite{shi2022learning} audio-visual self-supervised model,
and compute their $\ell_1$ norm on all samples in the FakeAVCeleb dataset \cite{khalid2021fakeavceleb}.
We perform the same analysis in the image domain using the DINOv3 \cite{simeoni2025dinov3} representations on the GenImage dataset \cite{zhu2023genimage}.
Figure~\ref{fig:main_l1_figure} shows the histograms of the norms for the two cases.
We observe that real samples and fake samples are very well separated by the $\ell_1$ norm, with the fake class being to the left (lower norm).

\begin{figure}[t]
\centering
\begin{subfigure}{0.49\columnwidth}
    \includegraphics[width=\linewidth]{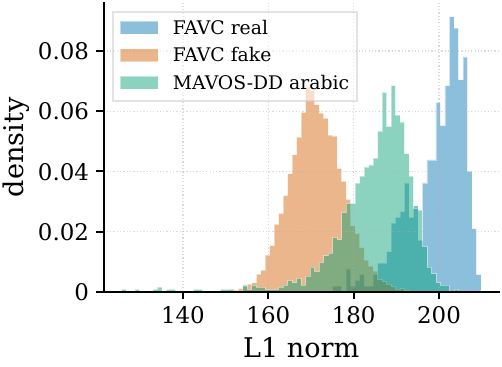} 
    \caption{Videos -- AV-HuBERT}
\end{subfigure}
\hfill
\begin{subfigure}{0.49\columnwidth}
    \includegraphics[width=\linewidth]{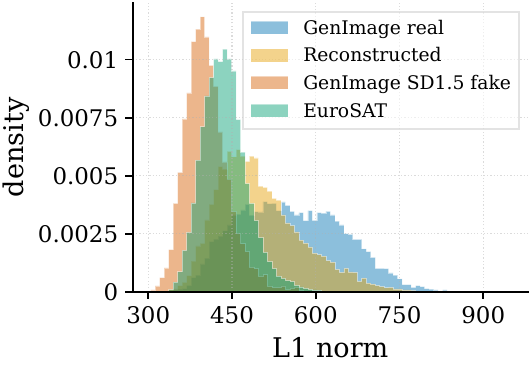} 
    \caption{Images -- DINOv3-7B}
    \label{fig:L1_norm_distribution_b}
\end{subfigure}

\caption{Histograms of $\ell_1$ feature norm. (a) AV-HuBERT features for real (blue) and fake (orange) videos from FakeAVCeleb, and real samples from MAVOS-DD Arabic (green). (b) DINOv3-7B features for real (blue), their reconstructed versions through SD1.5 (yellow) and Stable Diffusion v1.5 (SD1.5) fake (orange) images from GenImage, and EuroSAT satellite images (green). In both settings, real samples outside the encoder's pretraining distribution (green) have lower $\ell_1$ norms than in-distribution real samples (blue), but higher norms than fake samples (orange). The $\ell_1$ norm distribution of the reconstructed images falls between the two real categories, indicating that the effect of generative fingerprints is smaller than that of semantic distribution shift.}
\label{fig:L1_norm_distribution}
\end{figure}

\paragraph{Why does magnitude differ?}
While simple, this observation is surprising.
Why should such a basic statistic discriminate so well real from fake samples?
A similar behavior has been previously observed in classifiers:
feature norms tend to be smaller for unseen classes than for seen ones \citep{dhamija2018reducing, sun2021react, dietterich2022familiarity, Park_2023_ICCV, Yu_2023_CVPR}.
One reason is the \textit{familiarity hypothesis} \cite{dietterich2022familiarity}:
lower-magnitude features correspond to weaker activations, which in turn indicate absence of familiar features;
so, samples from unseen classes lack evidence present in seen classes.

A similar intuition applies in our case.
Foundation models are trained only on real data, making fake samples effectively out of distribution.
If the feature norm reflects familiarity, it should therefore distinguish real from fake samples as well.
But in what sense do real and fake samples differ?
The differences can stem from two factors:
semantic discrepancies or low-level generative fingerprints.
To isolate these effects, we design two experiments.
To evaluate the role of semantics, we consider real samples but from a distribution that differs from that used to pretrain the self-supervised encoder. %
To evaluate the role of low-level artifacts, we reconstruct real samples using a generative model;
this process injects generative fingerprints, while preserving the original semantics.
For video, we use MAVOS-DD Arabic \cite{croitoru-arXiv-2025} as the out-of-domain real data; this differs in identity, language, and phonetic content from VoxCeleb2 \cite{chung2018voxceleb2}, the main source for training AV-HuBERT.
For images, identifying truly out-of-distribution data is challenging because DINOv3 was trained on very diverse data from the internet.
We therefore use satellite imagery from EuroSAT \cite{helber2019eurosat} as a semantically distant domain.
For images, we also reconstruct real samples from GenImage with the SD1.5 autoencoder \cite{rombach2022high}, following the protocol of~\citet{ricker_2024_CVPR}. 

Figure~\ref{fig:L1_norm_distribution} shows the resulting distributions of the norms.
We observe that real samples from out-of-domain datasets (green histogram) have low norm, approaching those of fake samples.
We also observe that the re-generated real samples (yellow histogram on Figure~\ref{fig:L1_norm_distribution_b}) remain much closer to the real distribution than is to the fake one.
Taken together, these results suggest that the difference in norm is caused by the semantic shift rather the the low-level fingerprints.
The semantic discrepancies are likely introduced by the generative process, reflecting the inability of generative models to correctly reconstruct the real world.

\begin{figure}[t]
\centering

\begin{subfigure}{0.48\columnwidth}
    \centering
    \includegraphics[width=\linewidth]{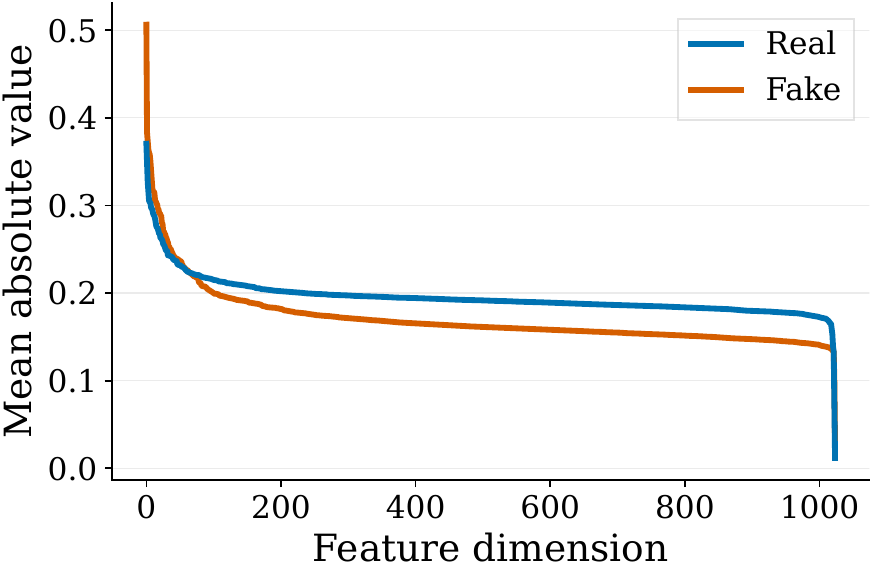}
    \caption{AV-HuBERT}
\end{subfigure}
\hfill
\begin{subfigure}{0.48\columnwidth}
    \centering
    \includegraphics[width=\linewidth]{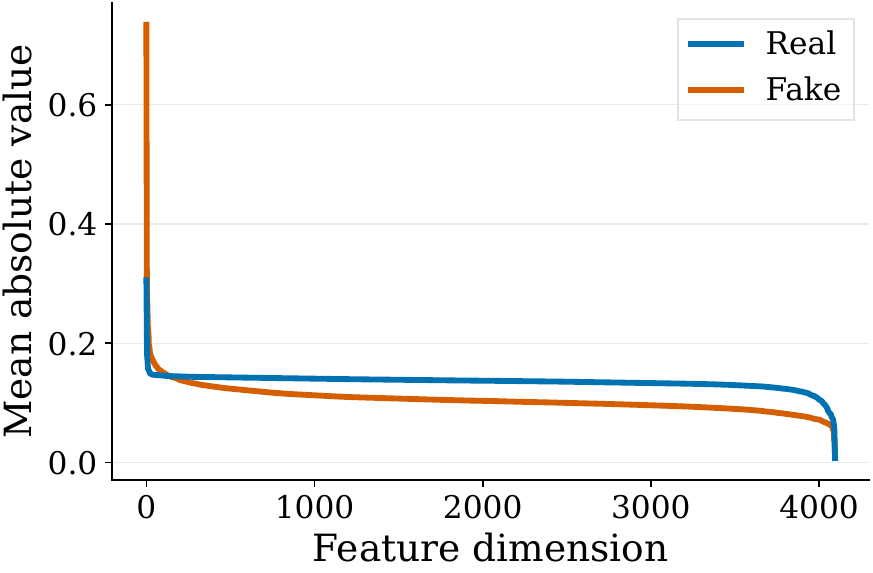}
    \caption{DINOv3-7B}
\end{subfigure}

\caption{Per-dimension mean absolute activations for (a) AV-HuBERT representations on FakeAVCeleb and (b) DINOv3-7B representations on GenImage.
The feature dimensions are sorted independently for each class.
Fake samples exhibit lower activation magnitudes across most dimensions, except for the highest-ranked activations, which are larger than those of real samples. 
This suggests that representations of fake samples are sparser than corresponding real ones. 
}
\label{fig:mean_features}
\end{figure}

\paragraph{Sparsity differs between real and fake samples.} 
The feature norm aggregates information across all dimensions.
To better understand the source of the observed norm differences, we take a closer look and analyze the activation of the individual feature dimensions. 
Specifically, we compute the mean absolute activation of each dimension separately for real and fake samples, sort the dimensions in descending order of their mean absolute activation for each class, and plot the resulting distributions.
Figure~\ref{fig:mean_features} shows the results.
At a high level, the differences appear across a large fraction of the representation dimensions rather than being confined to a small subset.
However, a closer inspection reveals an interesting pattern: the highest-ranked activations are slightly larger for fake representations than for real ones, whereas the remaining dimensions exhibit consistently lower activations.
This suggests that fake representations concentrate their energy in a relatively small subset of features, while real representations distribute it more evenly across the representation space.
In other words, fake representations are sparser, whereas real representations are more diffuse.

\paragraph{Detecting deepfakes using magnitude and sparsity.}
Motivated by the observations above, we propose to detect deepfakes using scores computed from self-supervised representations.
Our formulation is related to anomaly detection:
fake samples should produce scores that differ systematically from those of real samples.
Since the self-supervised foundation models are pretrained only on real media,
their representations \textit{implicitly} capture the distribution of real. Let $\mathbf{h}$ denote the model representation of a sample.
We consider two scores: magnitude and sparsity.
The magnitude is measured by the $\ell_1$ norm of the features:
\begin{equation}
\text{magnitude}(\mathbf{h}) = \|\mathbf{h}\|_1.
\end{equation}
The sparsity of the representation is estimated using the ratio between $\ell_1$ and $\ell_2$ norms \cite{hoyer2004nmf, lopes2013estimating}:
\begin{equation}
\text{sparsity}(\mathbf{h}) = \|\mathbf{h}\|_1/\|\mathbf{h}\|_2.
\end{equation}
This ratio quantifies how concentrated the representation is across feature dimensions while remaining invariant to global rescaling.
The quantity is numerically stable and has been shown to provide a lower bound on the true $\ell_0$ sparsity \cite{lopes2013estimating}.
To obtain a fakeness score  (larger values more likely to be fake), we use negative magnitude and negative norm ratio, as both are lower for fakes.
We refer to this method as NormFake, with its two variants denoted NormFake (magnitude) and NormFake (sparsity).

\begin{table*}[t]
\centering
\def\na{\color{gray}N/A}
\begin{tabularx}%
{\linewidth}{Xl cccccc c}
\toprule
Method & Fake train data & FAVC & AVLips & DFE & MAVOS-DD & MMDF & DFDC & Average \\
\midrule
\multicolumn{9}{l}{\textit{Fake-aware methods:}} \\

RealForensics & FF++ & 88.0 & 77.0 & 74.4 & 80.2 & 59.7 & 81.6 & 76.8 \\
AuViRe & AV1M & 87.9 & 72.4 & 63.8 & 62.0 & 77.8 & 63.8 & 71.3\\

\midrule
\multicolumn{9}{l}{\textit{Real-only methods:}} \\
AVAD & \na & 85.4 & 72.9 & 64.8 & 47.7 & 40.5 & 51.6 & 60.5\\ 
AVH-Align & \na & 93.1 & 86.3 & 49.8 & 50.3 & 53.2 & 34.2 & 61.2 \\

STALL & \na & 21.8 & 20.3 & 49.0 & 39.0 & 50.4 & 43.0 & 37.3 \\
FACTOR & \na & \underline{97.7} & \textbf{95.4} & 61.4 & 68.9 & 81.1 & 59.4 & 77.3 \\
SpeechForensics & \na & \textbf{98.5} & 92.7 & \textbf{75.8} & \textbf{81.3} & \textbf{92.6} & 59.6 & \textbf{83.4}\\ %

$\textbf{NormFake}$ (magnitude)  & \na & 97.2 & \underline{94.8} & 69.9 & \underline{79.0} & 76.5 & \textbf{65.1} & 80.4 \\  %

$\textbf{NormFake}$ (sparsity) & \na & 94.6 & 93.8 & \underline{72.8} & 78.8 & \underline{89.5} & \underline{64.7} & \underline{82.4} \\ %

\bottomrule
\end{tabularx}
\caption{Comparison with state-of-the-art video-level deepfake detection methods on six benchmarks (ROC-AUC \%). Our real-only approach, using only visual AV-HuBERT features, achieves competitive performance with prior multimodal methods, without relying on the audio modality. Best real-only results are shown in \textbf{bold}, and second-best results are \underline{underlined}.}

\label{tab:Video_SOTA}
\end{table*}

\section{Main results}

In this section, we evaluate the two variants of NormFake on video and image datasets and compare against prior work.

\subsection{Experimental Setup}

\paragraph{Video datasets.}
We evaluate on six audio-visual video datasets: FakeAVCeleb (FAVC) \cite{khalid2021fakeavceleb}, AVLips \cite{liu2024lips}, DeepfakeEval-2024 (DFE) \cite{Chandra_2026_CVPR}, MAVOS-DD \cite{croitoru-arXiv-2025}, MMDF \cite{Kim_2026_CVPR}, and DFDC \cite{dolhansky2020dfdc}.
We focus on audio-visual datasets because existing real-only video deepfake detection methods rely on audio-visual synchronization.
For completeness, we also report results on visual-only video datasets FaceForensics++~\cite{Rssler2019FaceForensicsLT}, DFD~\cite{dufour2019dfd}, DeeperForensics~\cite{Jiang2020DeeperForensics10AL}, and Celeb-DF-v2~\cite{Celeb_DF_cvpr20} in the supplementary material.

\paragraph{Video representations.}
We extract video representations using AV-HuBERT \cite{shi2022learning},
a self-supervised model that learns joint audio-visual representations.
We use the \texttt{self\_large\_vox\_433h} checkpoint,
pretrained on LRS3 \cite{afouras2018lrs3} and VoxCeleb2 \cite{chung2018voxceleb2}.
We extract visual features by inputting only the frames and masking the audio.
AV-HuBERT produces frame-level representations.
To obtain a per-video detection score,
we apply NormFake to each frame's feature vector and
then average these scores.

\paragraph{Image datasets.}
We use GenImage \cite{zhu2023genimage}, a million-scale benchmark that pairs real ImageNet images with fakes from eight generators (seven diffusion models plus BigGAN), each conditioned on the 1\,000 ImageNet classes.
The class-level alignment between real and fake images isolates detection performance from semantic content shifts.
We only consider its dedicated test set.

\paragraph{Image representations.}
We extract image representations using DINOv3 \cite{simeoni2025dinov3},
and use its largest available checkpoint, \texttt{DINOv3-7B}.
We represent each image using the embedding of the CLS token.
Unless stated otherwise, we use the final output embeddings.
However, as shown in the supplementary material,
embeddings from intermediate layers can outperform those from the final layer.

\paragraph{Evaluation metric.}
We report results in terms of the area under the receiver operating characteristic curve (ROC-AUC), which summarizes the trade-off between true positive rate and false positive rate across all decision thresholds.
As such, the metric is threshold-independent,
with higher values indicating better performance.
A score of 50\% corresponds to random chance, regardless of the class distribution.

\par\vspace{6pt}\noindent%
Datasets details, splits, preprocessing and more about the experimental setup are given in the supplementary material.

\subsection{Evaluation: Video deepfake detection}

To contextualize the results, we compare NormFake against prior methods.
We group these in fake-aware methods (trained using fake samples) and
real-only methods (trained solely on authentic media).
The fake-aware methods are
RealForensics \cite{haliassos2022leveraging},
AuViRe \cite{Koutlis_2026_WACV}.
The real-only methods are
AVAD \cite{feng2023self},
AVH-Align \cite{AVH-Align},
STALL \cite{ben2026stall},
FACTOR \cite{reiss2023detectingdeepfakesseeing}
SpeechForensics \cite{liang24speechforensics}.
For a fair comparison, we adapt STALL to our setting by using AV-HuBERT features and calibrating them on VoxCeleb2 (real samples).
NormFake is applied only in the visual domain. We evaluated audio representations but did not observe a comparable separation (see supplementary material).

The results are shown in Table~\ref{tab:Video_SOTA}.
Despite not having access to fake samples, NormFake outperforms fake-aware approaches on average.
Our approach also compares positively to most real-only approaches.
Only SpeechForensics \cite{liang24speechforensics} achieves a slightly higher average ROC-AUC,
with NormFake (sparsity) trailing by just one percentage point,
despite relying solely on the visual modality, whereas SpeechForensics leverages both audio and visual streams. This proves that NormFake, despite its simplicity, acts as a strong baseline for deepfake detection.

\begin{figure}[t]
\centering
\includegraphics[width=1\columnwidth]{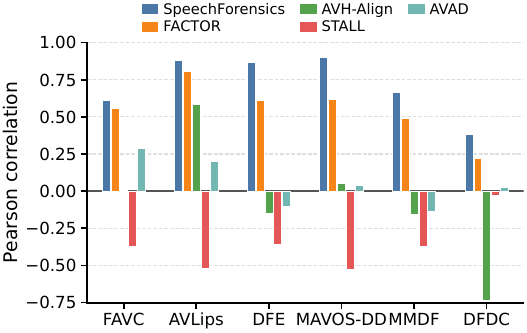} 
\caption{Pearson correlation of predictions between NormFake (magnitude) and other real-only methods.
Across datasets, NormFake exhibits strong correlation with SpeechForensics and FACTOR, while showing weak and even negative correlations with the remaining methods.}
\label{fig:bar_plot_score_correlation_SF_L1}
\end{figure}

To better understand NormFake, we examine how its decision boundary compares with that of other real-only methods.
To do so, we compute the Pearson correlation between the predictions of NormFake (magnitude) and those of each baseline.
Figure~\ref{fig:bar_plot_score_correlation_SF_L1} show that on most datasets NormFake is strongly correlated with SpeechForensics and FACTOR.
Both of these methods measure audio-visual synchronization using the cosine similarity between audio and visual AV-HuBERT representations.
It is therefore surprising that NormFake, which operates solely on visual representations, behaves so similarly.
One possible explanation is that the generative process simultaneously introduces visual artifacts and audio-visual synchronization inconsistencies, allowing the two approaches to capture complementary manifestations of the same underlying process.
The sparsity variant follows similar trends; see supplementary material.

\subsection{Evaluation: Image deepfake detection} 

For images, we compare NormFake with fake-aware methods FatFormer \cite{Liu_2024_CVPR}, AIDE \cite{yan2025sanity} and real-only methods, RIGID \cite{he2024rigid}, MINDER \cite{tsai2024minder}, AEROBLADE \cite{ricker_2024_CVPR}, Manifold Bias \cite{brokman2025manifold}, WarPAD \cite{choi2025warpad} and RDD \cite{choi2026debiased} in Table~\ref{tab:Image_SOTA}. 
To allow for a fair comparison,  for the two strongest baselines, RIGID and WarPAD, we report results with both the original backbone (DINOv2-L; \citealt{oquab2024dinov2learningrobustvisual}) and the more powerful DINOv3-7B backbone.
We do not directly compare to methods that require access to a generative model (RDD, AEROBLADE), as they rely on reconstruction error computed with a diffusion model.
As such, they assume knowledge of the generative process of the deepfakes they detect. 
Moreover, these methods' performance depends on the similarity between the diffusion model used for reconstruction and the one used to generate the test samples.

For NormFake (magnitude) we use output embeddings
while for the sparsity variant we chose the outputs of the penultimate block of the transformer. 
The latter method achieves state-of-the-art performance, particularly on continuous latent diffusion models such as Stable Diffusion 1.4/1.5 and Wukong, as well as the proprietary Midjourney model, but not on VQDM, which performs diffusion in a discrete quantized latent space. In contrast, prior methods achieve near-perfect performance on GAN-based and pixel-space diffusion models, outperforming NormFake. This suggests that the two families of methods exploit different cues in synthetic images. One possible explanation is that prior methods rely more heavily on pixel-level artifacts, whereas NormFake is more sensitive to the semantic deviations induced by continuous latent diffusion models. Moreover, for the two prior works evaluated on DINOv3-7B, we observe a slight drop in performance, suggesting that, unlike NormFake, they do not scale with the size of the backbone.
Details on layer choice are provided in the supplmentary material.

\begin{table*}[t]
\centering

\begin{tabularx}{\linewidth}{X cccccccc c}
\toprule
Method & ADM & BigGAN & GLIDE & Midjourney & SD1.4 & SD1.5 & VQDM & Wukong & Mean \\
\midrule
\multicolumn{10}{l}{\textit{Fake-aware methods:}} \\
FatFormer & 90.3 & 99.5 & 95.1 & 57.9 & 78.0 & 77.6 & 96.7 & 82.4 & 84.7 \\
AIDE & 92.1 & 92.0 & 98.7 & 95.9 & 100.0 & 100.0 & 96.5 & 100.0 & 96.9 \\
\midrule
\multicolumn{10}{l}{\textit{Real-only methods:}} \\
RIGID & \underline{87.4} & \underline{97.4} & \underline{95.2} & 77.8 & 68.2 & 68.2 & \underline{91.5} & 69.9 & 82.0 \\
MINDER & 76.8 & 68.1 & 58.2 & 45.0 & 60.7 & 59.6 & 88.2 & 67.6 & 65.5 \\

\textcolor{gray}{AEROBLADE} & \textcolor{gray}{85.1} & \textcolor{gray}{97.8} & \textcolor{gray}{98.9} & \textcolor{gray}{98.5} & \textcolor{gray}{97.6} & \textcolor{gray}{97.8} & \textcolor{gray}{72.1} & \textcolor{gray}{97.8} & \textcolor{gray}{93.2} \\
Manifold Bias & 68.1 & 92.7 & 82.2 & 47.6 & 66.3 & 65.9 & 88.0 & 64.1 & 71.9 \\
WaRPAD  & \textbf{98.6} & \textbf{99.8} & \textbf{99.1} & 81.0 & 94.0 & 93.6 & \textbf{98.1} & 92.4 & \underline{94.6} \\

\textcolor{gray}{RDD} & \textcolor{gray}{92.6} & \textcolor{gray}{99.7} & \textcolor{gray}{99.6} & \textcolor{gray}{98.3} & \textcolor{gray}{100.0} & \textcolor{gray}{99.9} & \textcolor{gray}{94.6} & \textcolor{gray}{99.9} & \textcolor{gray}{98.1} \\

\midrule

RIGID$^\dagger$       & 84.5 & 84.2 & 93.4 & 76.9 & 76.5 & 76.8 & 76.0 & 76.0 & 80.5  \\
WaRPAD$^\dagger$      & 93.3 & 99.6 & 92.6 & 65.7 & 77.6 & 77.1 & 93.0 & 70.1 & 83.6 \\

$\textbf{NormFake}$ (magnitude) & 75.1 & 80.3 & 91.1 & \underline{91.1} & \underline{94.7} & \underline{94.2} & 75.9 & \underline{93.0} & 
86.9 \\

$\textbf{NormFake}$ (sparsity) & 86.7 & 96.3 & 96.7 & \textbf{94.2} & \textbf{98.0} & \textbf{97.5} & 93.8 & \textbf{97.7} & \textbf{95.1} \\

\bottomrule
\end{tabularx}
\caption{%
Comparison with state-of-the-art image deepfake detection methods on GenImage (ROC-AUC \%).
Gray: methods that require access to generative models.
$^\dagger$:
methods that use DINOv3-7B features (for a fair comparison with NormFake).
For NormFake, we report the best-performing layer for each variant.
The sparsity variant achieves the highest average performance.
Best real-only results are shown in \textbf{bold}, and second-best results are \underline{underlined}.
}
\label{tab:Image_SOTA}
\end{table*}

\begin{figure}[t]
\centering
\includegraphics[width=\linewidth]{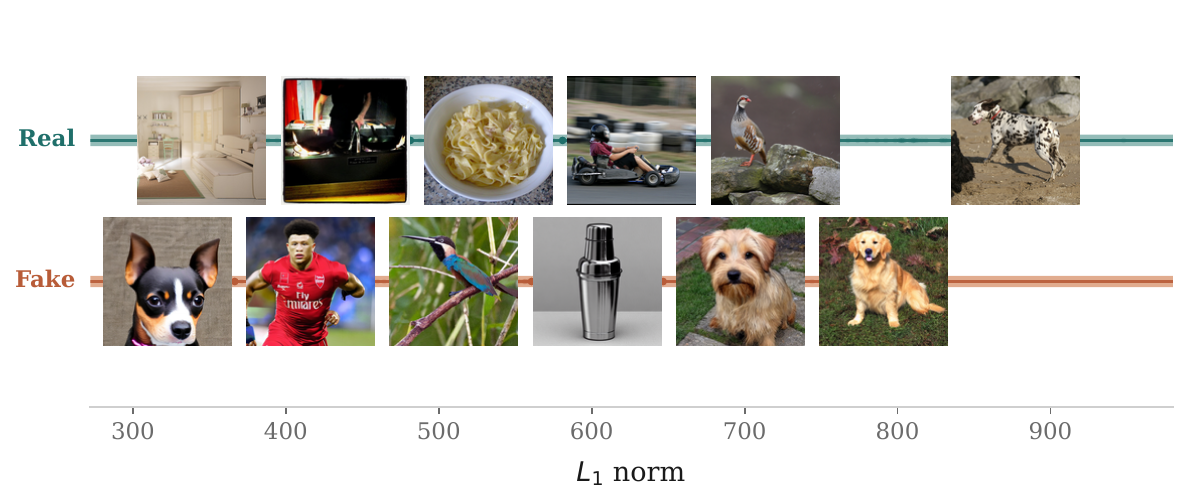}
\caption{Qualitative examples of DINOv3-7B features with their associated $\ell_1$ norm for GenImage reals and SD1.5 fakes.}
\label{fig:qualitative_examples}
\end{figure}

\paragraph{Qualitative examples.} In Figure~\ref{fig:qualitative_examples} we provide examples of real and fake images from GenImage (SD1.5) with both low and high $\ell_1$ norms, using DINOv3-7B representations. Images with lower norms exhibit a more pronounced synthetic style, whereas higher-norm images contain fewer generative artifacts. One interesting observation is that real images with low $\ell_1$ norms often resemble cluttered scenes or poorly lit photos, often harder to identify as real, while those with higher norms tend to depict cleaner, object-centric imagery. This is in line with our hypothesis that the $\ell_1$ norm of pretrained features distinguishes between real and fake semantics.

\begin{figure}[t]
\centering
\includegraphics[width=\columnwidth]{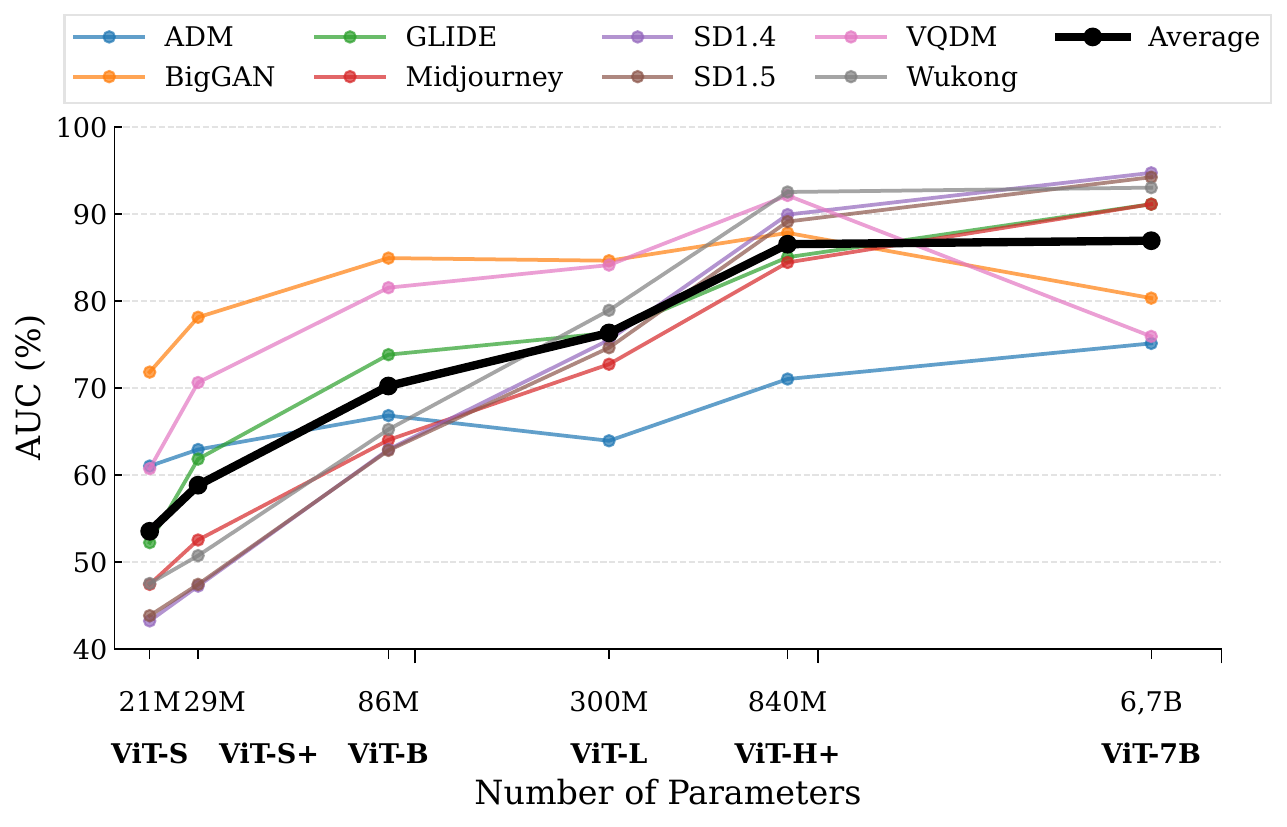} 
\caption{ROC-AUC of NormFake (magnitude) using DINOv3 features across different model scales on GenImage. Performance increases steadily with model size.} %
\label{fig:ablation_network_size}
\end{figure}

\section{Further analyses}

We now investigate the impact of model scale, backbone, and feature extraction layers.

\paragraph{Model scale.}
To evaluate the effect of model scale, %
we consider six transformer variants from the DINOv3 family, 
ViT-S, ViT-S+, ViT-B, ViT-L, ViT-H+, and ViT-7B.
The results are shown in Figure~\ref{fig:ablation_network_size} for the eight GenImage datasets,
together with their average performance (black line).
We observe that average performance improves as model size increases, with gains gradually saturating at the ViT-H+ variant, whose performance is comparable to that of the substantially larger ViT-7B model. This is partially driven by a decrease in performance on BigGAN and VQDM. One possible explanation is that larger foundation models learn increasingly semantic representations, while smaller backbones remain relatively more sensitive to low-level artifacts. We hypothesize these artifacts are more pronounced in BigGAN and VQDM due to their GAN-based upsampling and discrete-latent quantization.

These results suggest that larger foundation model representations yield stronger performance, with the model scale serving as a reliable proxy for representation quality.

\paragraph{Backbones.}
We consider five backbone families: BEiT \cite{bao2022beit}, OpenCLIP \cite{cherti2023reproducible}, SigLIP 2 \cite{tschannen2025siglip2multilingualvisionlanguage}, PE-Core \cite{bolya2025PerceptionEncoder} and DINOv3.
For each backbone we use the largest available variant.
As shown in Table~\ref{tab:ablation_image_backbone_avg}, NormFake (magnitude) presents a clear scaling trend with foundation model size, across the backbones. PE-CORE-G14 achieves the highest performance of 87.7\% ROC-AUC, closely followed by DINOv3-7B.
In contrast, sparsity performs near chance, except with DINOv3-7B, scoring comparable with the magnitude variant. This suggests that feature sparsity is less reliable than magnitude, as fewer models exhibit discriminative sparsity patterns. 

We further evaluate the impact of the backbone on the video setup in Table ~\ref{tab:ablation_image_backbone_avg}.
We consider RAVEn \cite{haliassos2025raven} and BRAVEn \cite{haliassos2024braven} visual-only representations and evaluate on FakeAVCeleb.
Unlike the image setting, the sparsity variant also achieves competitive performance across all evaluated video backbones.
In this setting, the number of parameters is similar across backbones.

\begin{table}[t]
\centering
\begin{tabular}{lr rr}
\toprule
& & \multicolumn{2}{c}{NormFake} \\
\cmidrule(lr){3-4}
Network & No. Params. & Magnitude & Sparsity \\
\midrule
\multicolumn{4}{l}{\textit{Image-based models $\,\cdot\,$ Evaluated on GenImage}} \\
BEiT-L & 303M & 57.3 & 59.0\\
OpenCLIP-G/14 & 1.01B & 61.9 & 59.4\\
SigLIP2-Giant & 1.16B & 73.5 & 52.7\\
PE-Core-G14 & 1.88B & \bf 87.7 & 53.7 \\
DINOv3-7B  & 6.72B & 86.9 & \bf 83.2\\
\midrule
\multicolumn{4}{l}{\textit{Video-based models $\,\cdot\,$ Evaluted on FakeAVCeleb}} \\
RAVEn & 339M & 95.7 & 87.2 \\
BRAVEn & 339M & 95.0 & 87.5 \\
AV-HuBERT & 324M & \bf 97.2 & \bf 94.6\\
\bottomrule
\end{tabular}
\caption{Average performance (ROC-AUC\%) of NormFake across the GenImage generators for image encoders and FakeAVCeleb for video encoders using multiple foundation model backbones. The approach provides consistently strong performance across all evaluated backbones.}
\label{tab:ablation_image_backbone_avg}
\end{table}

\begin{figure}[t]
    \centering

    \begin{subfigure}{0.9\columnwidth}
        \centering
        \includegraphics[width=\linewidth]{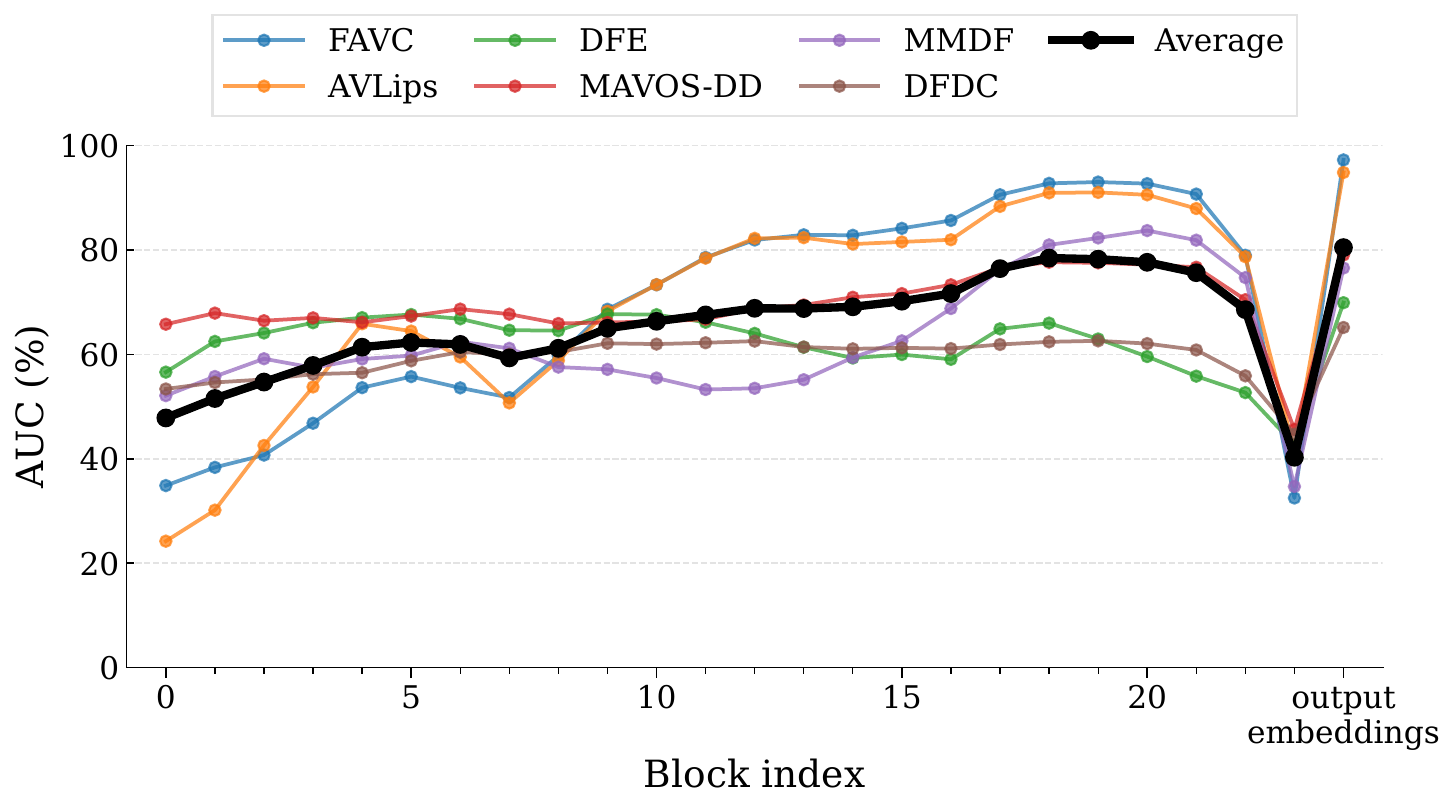}
        \caption{AV-HuBERT}
    \end{subfigure}
    \hfill
    \begin{subfigure}{0.9\columnwidth}
        \centering
        \includegraphics[width=\linewidth]{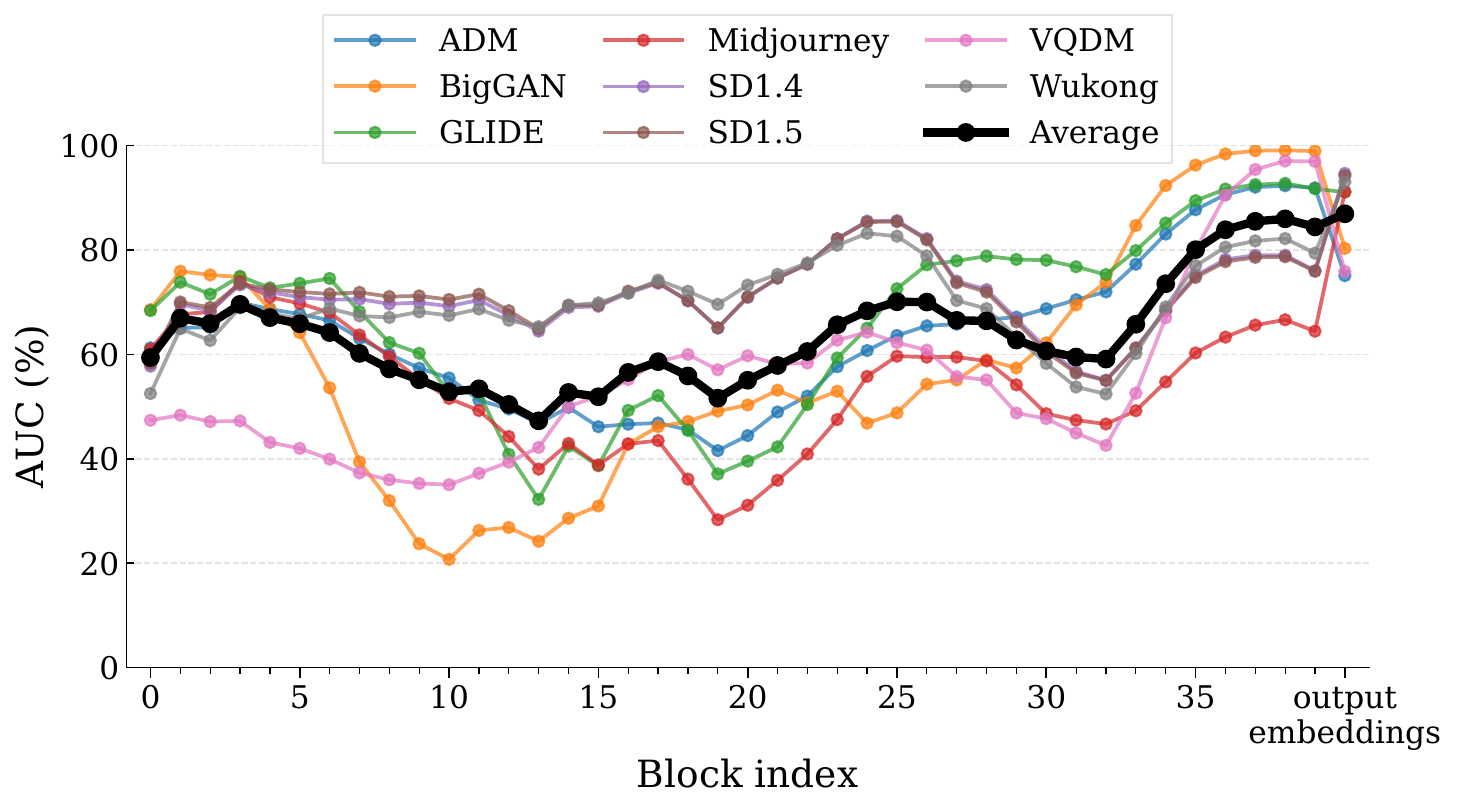}
        \caption{DINOv3-7B}
    \end{subfigure}

    \caption{Per transformer block and output embeddings ROC-AUC\% of NormFake (magnitude) using (a) AV-HuBERT features and (b) DINOv3-7B features.
    Later layers provide the best performance, while the behavior of earlier layers differs between models.}
    \label{fig:per_layer_roc_auc_norm}
\end{figure}

\paragraph{Per-layer performance.}
Lastly, we analyze how the $\ell_1$-norm separability between real and fake samples evolves across network layers.
Figure~\ref{fig:per_layer_roc_auc_norm} reports the ROC-AUC obtained from features at the end of each transformer block, as well as from the final output embeddings, obtained by applying the final LayerNorm to the output of the last transformer block. We do so for AV-HuBERT on the video benchmarks and DINOv3-7B on GenImage. 

For AV-HuBERT, the average performance increases steadily with network depth. Interestingly, the output of the final transformer block exhibits a pronounced drop in performance, whereas the network's final output representation recovers to the highest ROC-AUC. This suggests that the final LayerNorm substantially enhances separability between real and fake representations. 
One possible explanation is that feature norms naturally increase with depth, reducing the relative discrepancy between reals and fakes. The final LayerNorm rescales the representations, restoring seperability.

In contrast, DINOv3 displays a noisier progression throughout the first three quarters of the network. Nevertheless, from the 32nd transformer block, performance improves consistently across all generators, culminating in the highest ROC-AUC for the final CLS embedding. Notably, even the earliest layers achieve around 70\% ROC-AUC, indicating that early visual representations already contain cues that distinguish generated images from real ones.

Both architectures achieve the strongest separation in the deepest representations, suggesting that feature-magnitude discrepancies are amplified as representations become more semantic, while remaining partially detectable in earlier layers.
However, sparsity can favour late over final layers regardless of the GenImage generator; see supplementary material.

\section{Limitations}
\paragraph{Over-reliance on pretrained networks.}
NormFake relies entirely on representations extracted from pretrained foundation models. %
As a result, its performance depends on both the representation capacity of the backbone and its pretraining data diversity.
For example, if a particular real domain is underrepresented,
then its samples may be assigned unusually low feature norms and consequently flagged as anomalous.

\paragraph{Pretraining data contamination.} 
As AI-generated content becomes increasingly prevalent online, future self-supervised foundation models may be pretrained on increasingly contaminated data.
This weakens the assumption that foundation models learn a representation shaped predominantly by real data,
potentially reducing the distinction between real and fake samples on which NormFake relies.

\section{Conclusions and future work}
We identified a consistent behavior of pretrained foundation models: fake media systematically produce lower-magnitude features than real media, across images and videos, diverse backbones, and a wide range of generators and manipulation types. 
Building on this observation, we show that simple feature statistics provide a strong deepfake detection baseline matching or outperforming considerably more complex methods. Our analysis further indicates that this phenomenon is driven primarily by the semantic shift introduced by the generative process, rather than low-level fingerprints, and that its discriminative strength consistently increases with model scale. Together, these findings suggest that detecting fake content can be an implicit byproduct of learning a faithful representation of real-world data, motivating simple and interpretable complements to specialized deepfake detectors.
\paragraph{Future work.}
We believe that our work opens several promising research directions. One is to incorporate representation statistics into trainable deepfake detectors through initialization priors or regularization terms. Another direction is to explore more informative statistics and pretraining objectives to discover stronger deepfake detection signals.

\bibliography{aaai2027}

\include{Supp_arxiv}

\end{document}

%% file: Supp_arxiv.tex
\begin{center}
    {\Large\bfseries Supplementary Material}
\end{center}

\section{Datasets}

In this section, we briefly describe the datasets used for evaluation, along with the implementation details.

\subsection{Video datasets} 

\paragraph{FakeAVCeleb (FAVC)~\cite{khalid2021fakeavceleb}} FakeAVCeleb is an audio-visual deepfake benchmark containing subjects from diverse ethnic backgrounds. It combines multiple visual manipulations, including face swapping (FaceSwap~\cite{Korshunova2017FastFaceSwap}, FSGAN~\cite{Nirkin2019FSGAN}) and facial reenactment (Wav2Lip~\cite{Prajwal2020ALip}), with synthetic audio generated through voice cloning (SV2TTS~\cite{Jia2018TransferLearningSpeakerVerification}) and lip synchronization. 

\paragraph{AVLips~\cite{liu2024lips}.} AVLips focuses on lip synchronization manipulations while preserving the remainder of the face and identity, using four generators: MakeItTalk~\cite{zhou2020makelttalk}, Wav2Lip~\cite{Prajwal2020ALip}, TalkLip~\cite{wang2023seeing}, SadTalker~\cite{Zhang2023SadTalker}.

\paragraph{DeepfakeEval-2024 (DFE)~\cite{Chandra_2026_CVPR}.} DeepfakeEval-2024 is a large-scale multimodal benchmark of in-the-wild deepfakes collected from social media in 2024. Samples are manually labeled and the underlying generators are typically unknown.

\paragraph{MAVOS-DD~\cite{croitoru-arXiv-2025}.} MAVOS-DD is a multilingual open-set benchmark using seven generators: talking-head (EchoMimic~\cite{Chen2024EchoMimic}, Memo~\cite{Zheng2026MEMO}, Sonic~\cite{Ji2025SONIC}), expression transfer (LivePortrait~\cite{Guo2024LivePortrait}), face swap (Inswapper\footnote{https://github.com/deepinsight/insightface}, HifiFace~\cite{Wang2021HifiFace}, Roop\footnote{https://github.com/s0md3v/roop}).

\paragraph{MMDF~\cite{Kim_2026_CVPR}.} MMDF evaluates generalization to recent portrait animation models. Its test set contains manipulations generated by HunyuanAvatar~\cite{Chen2025HunyuanVideoAvatar}, MegActor-$\Sigma$~\cite{Yang2025MegaActorSigma}, and AniPortrait~\cite{Wei2024AniPortrait}, which synthesize realistic talking-head videos from source images and driving signals.

\paragraph{DFDC~\cite{dolhansky2020dfdc}.} DFDC  is a widely used benchmark for deepfake detection. It contains videos generated using several manipulation pipelines, including DFAE~\cite{perov2020deepfacelab}, an autoencoder-based face-swapping method; MM/NN~\cite{huang2012facial}, a classical landmark-based face-swapping approach that warps and blends faces without using deep generative models; and three neural generation methods: Neural Talking Heads~\cite{zakharov2019few}, FSGAN~\cite{Nirkin2019FSGAN}, and StyleGAN~\cite{karras2019style}.

\paragraph{Implementation.} For FAVC and AVLips we consider the whole dataset. For MAVOS-DD, DFE and MMDF, we use their designated test sets. For DFDC, due to its large size, we evaluate only on its last two partitions (48 and 49). 

Following the protocol of~\citet{ssr-dfd}, we preprocess DFE to retain only samples that satisfy: 
\begin{enumerate*}[label=(\roman*)]
    \item each video segment that has an associated audio stream;
    \item a video segment should contain a single speaking face that was tracked in every frame;
    \item the identified face is larger than 100px×100px
    \item the duration of audio-video segment is between 3 and 60 seconds. If a video exceeds 60 seconds, we split it into chunks of at most 60 seconds.
\end{enumerate*}
Samples with an \textit{unknown} class label are also discarded.

Due to the nature of MAVOS-DD, which includes TED Talks and news segments that do not always show the person speaking, we apply the same preprocessing to this dataset. Furthermore, we evaluate only on English videos from the MAVOS-DD test set to avoid introducing semantic distribution shifts within the real domain.

As AV-HuBERT relies on cropped faces, we retain only videos for which its preprocessing pipeline successfully detects a face. We report the number of resulting videos after preprocessing for each dataset in Table~\ref{tab:dataset_preprocessing}.

\begin{table}[h]
\centering
\begin{tabularx}{\linewidth}{Xrr}
\toprule
Dataset & Original & Final \\
\midrule
FakeAVCeleb & 21566 & 21555 \\ 
AVLips & 7557 & 7557 \\
Deepfake-Eval-2024 & 815 & 577 \\
MAVOS-DD & 7951 & 7011 \\
MMDF & 4884 & 4477 \\
DFDC & 5597 & 5453 \\
\bottomrule
\end{tabularx}
\caption{Number of evaluation samples before and after preprocessing for each dataset.}
\label{tab:dataset_preprocessing}
\end{table}

\subsection{Image datasets}
\paragraph{GenImage~\cite{zhu2023genimage}.} GenImage is a million-scale benchmark for AI-generated image detection. Real images are drawn from ImageNet, and fake images are generated by eight generators spanning GANs and diffusion models: BigGAN~\cite{Brock2019BigGAN}, ADM~\cite{Dhariwal2021DiffusionModelsBeatGANs}, GLIDE~\cite{Nichol2021GLIDETP}, VQDM~\cite{Gu2022VQDiffusion}, Wukong~\cite{Wukong2022}, Midjourney~\cite{Midjourney2022}, and Stable Diffusion v1.4/v1.5~\cite{rombach2022latent}. Each generator subset has its own dedicated, non-overlapping real images, rather than a shared real pool, so real and fake images stay balanced within each generator subset. The evaluation is performed on each real–generator subset.

\paragraph{Implementation.} We evaluate on the test sets that contains 6\,000 real images and 6\,000 fake images per generator. One exception to this is SD1.5, which has 8\,000 samples for each class.

\section{Backbones}
\label{sec:supp-backbones}

\subsection{Video backbones}

\textbf{AV-HuBERT~\cite{shi2022learning}.} This network is trained via masked multimodal cluster prediction: masked audio and visual streams predict discrete cluster targets obtained from $k$-means on prior-iteration features, with modality dropout preventing reliance on a single stream.

\textbf{RAVEn~\cite{haliassos2025raven}.} Replaces AV-HuBERT's offline clustering with cross-modal self-distillation: each modality's encoder predicts targets from a momentum copy of the other modality's encoder.

\textbf{BRAVEn~\cite{haliassos2024braven}.} Extends RAVEn with asymmetric augmentations, more prediction targets, and a revised momentum schedule for stronger visual-only (lip-reading) representations.

\subsection{Image backbones}
\textbf{DINOv3~\cite{simeoni2025dinov3}.} DINOv3 is a self-supervised vision transformer trained without labels or text, using a self-distillation objective where a student network matches the output of a teacher on different augmented views of the same image. It is trained on diverse web-scale data.

\textbf{BEiT~\cite{bao2022beit}.} BEiT pretrains a vision transformer with a BERT-style masked-image-modeling objective: images are tokenized into discrete visual tokens, and the model learns to predict masked patches from context. Unlike contrastive methods, its pretraining signal is purely reconstructive and does not involve text supervision.

\textbf{OpenCLIP~\cite{ilharco_gabriel_2021_5143773}.} This model is an open-source reproduction of CLIP~\cite{radford2021learning}, trained with a contrastive image-text objective on large-scale web data so that matching image-text pairs are pulled together in a shared embedding space.

\textbf{SigLIP 2~\cite{tschannen2025siglip2multilingualvisionlanguage}.} SigLIP builds on CLIP-style contrastive pretraining but replaces the softmax contrastive loss with a sigmoid loss, improving training efficiency, and incorporates multilingual data and auxiliary objectives to strengthen semantic understanding and dense feature quality.

\textbf{PE-Core~\cite{bolya2025PerceptionEncoder}.} Perception Encoder is a contrastively-trained general-purpose visual encoder designed to transfer well across a broad range of downstream vision tasks, motivated by the observation that the strongest embeddings are often found in intermediate rather than final layers. It introduces language alignment, which fine-tunes intermediate features for multimodal language modeling, and spatial alignment, which distills intermediate features into dense, spatially-consistent representations.

\section{Additional experiments}

\subsection{Results on visual-only video}

Table~\ref{tab:ablation_video_only} reports results of NormFake on visual-only video benchmarks,
which do not include audio:
FaceForensics++ (FF++)~\cite{Rssler2019FaceForensicsLT},
DFD~\cite{dufour2019dfd},
DeeperForensics (DFo)~\cite{Jiang2020DeeperForensics10AL}, and
Celeb-DF-v2 (CDFv2)~\cite{Celeb_DF_cvpr20}.
We compare with the closest real-only prior work, namely STALL,
which we calibrate using visual AV-HuBERT features, as described in the main paper.
Additionally, we also report results for a fake-aware method, RealForensics, that is trained on FaceForensics++. 

We observe that NormFake's results are quite moderate, lagging behind RealForensics, but above the sub-random chance performance of STALL.
This strengthens our claim that NormFake acts as a strong non fake-aware baseline for deepfake detection.
The results for RealForensics are taken from~\citet{Kim_2025_ICCV}.

\begin{table}[h]
\centering
\setlength{\tabcolsep}{5pt}
\begin{tabularx}{\linewidth}{Xrrrr}
\toprule
Method & FF++ & DFD & DFo. & CDFv2 \\
\midrule
\multicolumn{5}{l}{\textit{Fake-aware methods:}} \\
RealForensics & \textit{Train} & 82.2 & 99.3 & 86.9 \\  
\midrule
\multicolumn{5}{l}{\textit{Real-only methods:}} \\
STALL & 32.7 & 41.4 & 47.1 & 42.2 \\
\textbf{NormFake} (magnitude)
& 85.9
& 62.2
& 79.0
& 64.6 \\

\textbf{NormFake} (sparsity)
& 78.1
& 64.2
& 82.4
& 62.6 \\

\bottomrule
\end{tabularx}
\caption{Performance (ROC-AUC, \%) on visual-only video datasets, without access to the audio stream. \textit{Train} indicates that FF++ is the training dataset used by RealForensics and is therefore not evaluated.
NormFake achieves competitive performance, trailing the FF++-trained RealForensics,
while substantially outperforming STALL.}
\label{tab:ablation_video_only}
\end{table}

\begin{figure}[!t]
    \centering

    \begin{subfigure}{1\linewidth}
        \centering
        \includegraphics[width=\linewidth]{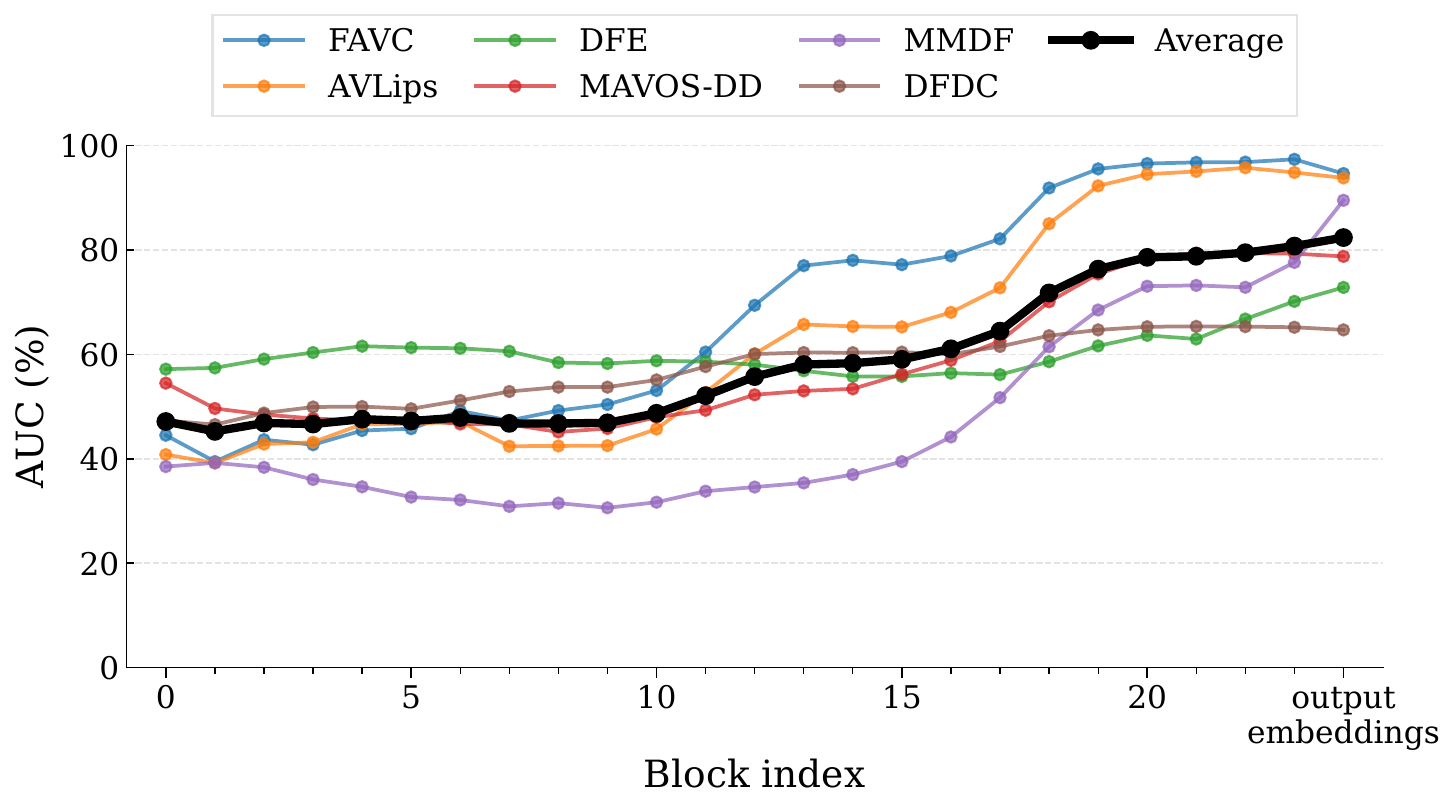}
        \caption{AV-HuBERT}
        \label{fig:per_layer_l2}
    \end{subfigure}
    \hfill
    \begin{subfigure}{1\linewidth}
        \centering
        \includegraphics[width=\linewidth]{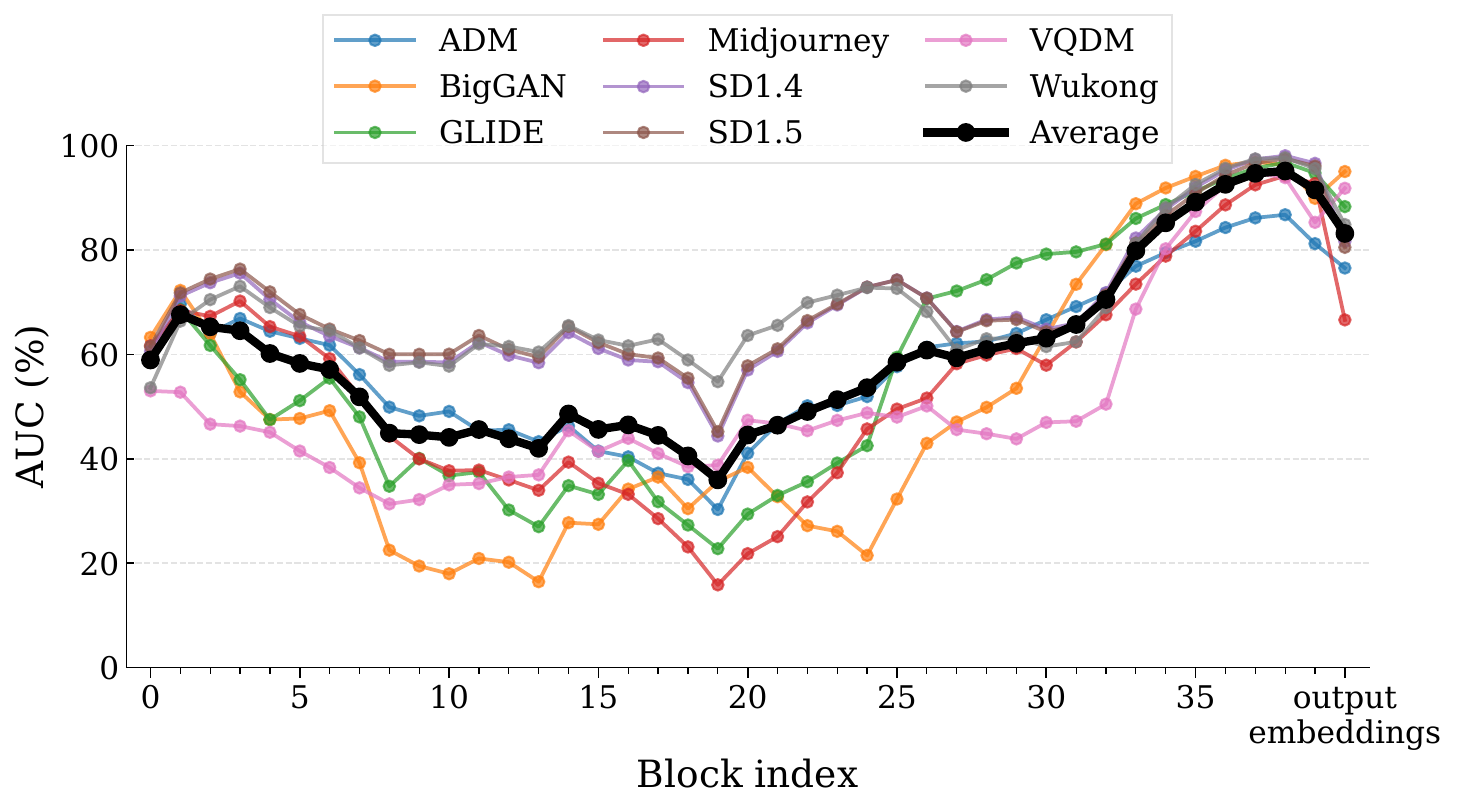}
        \caption{DINOv3-7B}
        \label{fig:per_layer_l1}
    \end{subfigure}

    \caption{Per transformer block and output embeddings ROC-AUC \% of NormFake (sparsity) using (a) AV-HuBERT features on videos and (b) DINOv3-7B features on images.
    Later layers provide the best performance, while the behavior of earlier layers differs between models. Specifically, for the video setting, AV-HuBERT achieves its best performance at the last layer, while in the image setting, DINOv3-7B achieves its best performance at the end of the penultimate transformer block. The latter is valid for all generators.}
    \label{fig:per_layer_image_performance}
\end{figure}

\subsection{Per-layer analysis of NormFake (sparsity)}

Figure~\ref{fig:per_layer_image_performance} shows the performance of NormFake (sparsity) when using features from the end of each transformer block or final embeddings.
We observe that the final layer is not always best.
On DINOv3-7B, in Figure~\ref{fig:per_layer_l1}, the best layer is actually the penultimate transformer block.
Notably, this trend is consistent across all generators in GenImage, suggesting that this layer captures a representation that is broadly favorable for image-level deepfake detection rather than being generator-specific.
This observation indicates that, for some backbones, intermediate representations may be more informative than the final embedding
A similar phenomenon has been reported in the out-of-domain detection setups, where penultimate-layer representations of classifiers have been shown to provide stronger separation between in-distribution and out-of-distribution samples than the final layer~\cite{Yu_2023_CVPR}.

\subsection {Model correlation for NormFake (sparsity)}

\begin{figure}[ht]
\centering
\includegraphics[width=1\columnwidth]{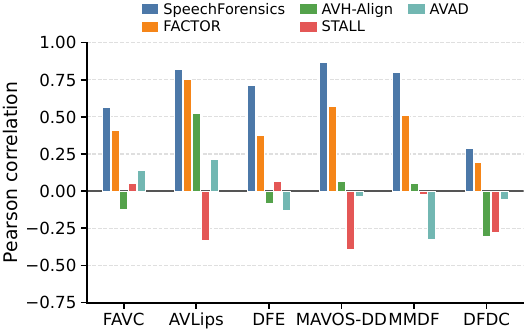} 
\caption{Pearson correlation between the sample-level predictions of real-only deepfake detectors and NormFake (sparsity).
Across datasets, SpeechForensics and FACTOR are consistently highly correlated with NormFake, whereas the remaining methods exhibit weak or even negative correlations.}
\label{fig:bar_plot_score_correlation_L1_L2}
\end{figure}

We compare the NormFake (sparsity) predictions with those produced by other real-only deepfake detectors by computing the Pearson correlation coefficient between those.
The results are shown in Figure~\ref{fig:bar_plot_score_correlation_L1_L2}.
We observe a similar trend to the results in the paper, which used the NormFake (magnitude) variant:
SpeechForensics and FACTOR exhibit consistently high positive correlations with NormFake,
whereas the remaining real-only methods show weak or even negative correlations.
These findings further support our hypothesis that SpeechForensics, FACTOR, and NormFake capture complementary manifestations of the same underlying process.

\subsection{NormFake on other modalities}

\paragraph{Audio.}
We evaluate how discriminative is NormFake for audio-based deepfake detection.
We do so by using audio-only features extracted from AV-HuBERT, RAVEn, and BRAVEn on the FakeVideo-FakeAudio (FVFA) and RealVideo-FakeAudio (RVFA) subsets of FakeAVCeleb.
For AV-HuBERT, we mask the visual branch and use only the audio waveform as input.
Real samples are taken from the RealVideo-RealAudio split.
The results are reported in Table~\ref{tab:Audio_FAVC}.
Unlike the visual setting, we do not observe a consistent separation between real and fake samples across backbones.
AV-HuBERT achieves good performance only on samples containing both visual and audio manipulations, and only with the magnitude variant.
In contrast, the same variant performs reasonably well with RAVEn and BRAVEn on both splits, whereas the sparsity variant consistently performs below chance across all three backbones.
Overall, while these results suggest that NormFake may capture useful cues from audio representations, they indicate that the phenomenon is substantially weaker than in the visual domain and warrants further investigation, particularly on dedicated audio-only deepfake benchmarks.

\begin{table}[h]
\centering
\def\na{\color{gray}N/A}
\begin{tabularx}%
{\linewidth}{Xl rr}
\toprule
Backbone & Variant & FVFA & RVFA \\
\midrule
\multirow{2}{*}{AV-HuBERT} & magnitude & 78.2 & 38.7 \\ %
& sparsity  & 42.4 & 34.1 \\  %
\midrule
\multirow{2}{*}{BRAVEn}    & magnitude & 60.5 & 74.2 \\ %
& sparsity  & 17.2 & 32.7 \\  %
\midrule
\multirow{2}{*}{RAVEn}     & magnitude & 62.3 & 80.6 \\ %
& sparsity  & 15.8 & 43.2 \\  %
\bottomrule
\end{tabularx}
\caption{Performance of NormFake using audio-only features from AV-HuBERT, RAVEn, and BRAVEn on the FakeAVCeleb splits containing manipulated audio. Overall, the magnitude variant yields moderate performance on most backbones and splits, whereas the sparsity variant remains close to chance.}

\label{tab:Audio_FAVC}
\end{table}

\paragraph{Multimodal.} We also evaluate NormFake using joint audio-visual representations extracted from AV-HuBERT, where both the audio waveform and visual frames are provided as input. Table~\ref{tab:Multimodal_FAVC} reports the results for each FakeAVCeleb split, together with those obtained using visual-only and audio-only representations for comparison. NormFake (magnitude), i.e.\, the $\ell_1$ norm, achieves strong performance on the FakeVideo-FakeAudio (FVFA) and RealVideo-FakeAudio (RVFA) splits, but drops to near chance on the FakeVideo-RealAudio (FVRA) split, where only the visual stream is manipulated. In contrast, the $\ell_1/\ell_2$ variant remains close to chance across all splits. Compared with the unimodal settings, visual-only representations perform best whenever the video is manipulated, whereas only the multimodal representation performs well when only the audio is manipulated, unlike audio features which perform above chance only on FakeVideo-FakeAudio (FVFA). These inconsistencies make it unclear what information is encoded by each representation, with only the visual-only features behaving as expected.

\begin{table}[h]
\centering
\begin{tabular}{llcccc}
\toprule
Features & Variant & FVFA & RVFA & FVRA & All \\
\midrule
\multirow{2}{*}{Multimodal}
& $\ell_1$ & 84.0 & 88.5 & 53.0 & 69.8 \\
& $\ell_1 / \ell_2$  & 49.6 & 47.4 & 54.2 & 51.7 \\
\midrule
Visual
& $\ell_1$  & 99.6 & 54.6 & 96.8 & 97.2 \\
\midrule
Audio
& $\ell_1$  & 78.2 & 38.7 & 49.2 & 63.9 \\
\bottomrule
\end{tabular}
\caption{Performance of NormFake using AV-HuBERT multimodal, visual-only, and audio-only representations on the FakeAVCeleb splits. Visual-only representations achieve the highest performance, while multimodal and audio-only representations are effective primarily on splits containing audio manipulations.}
\label{tab:Multimodal_FAVC}
\end{table}

\subsection{Qualitative examples}

\begin{figure*}[t]
    \centering
    \includegraphics[width=0.85\linewidth]{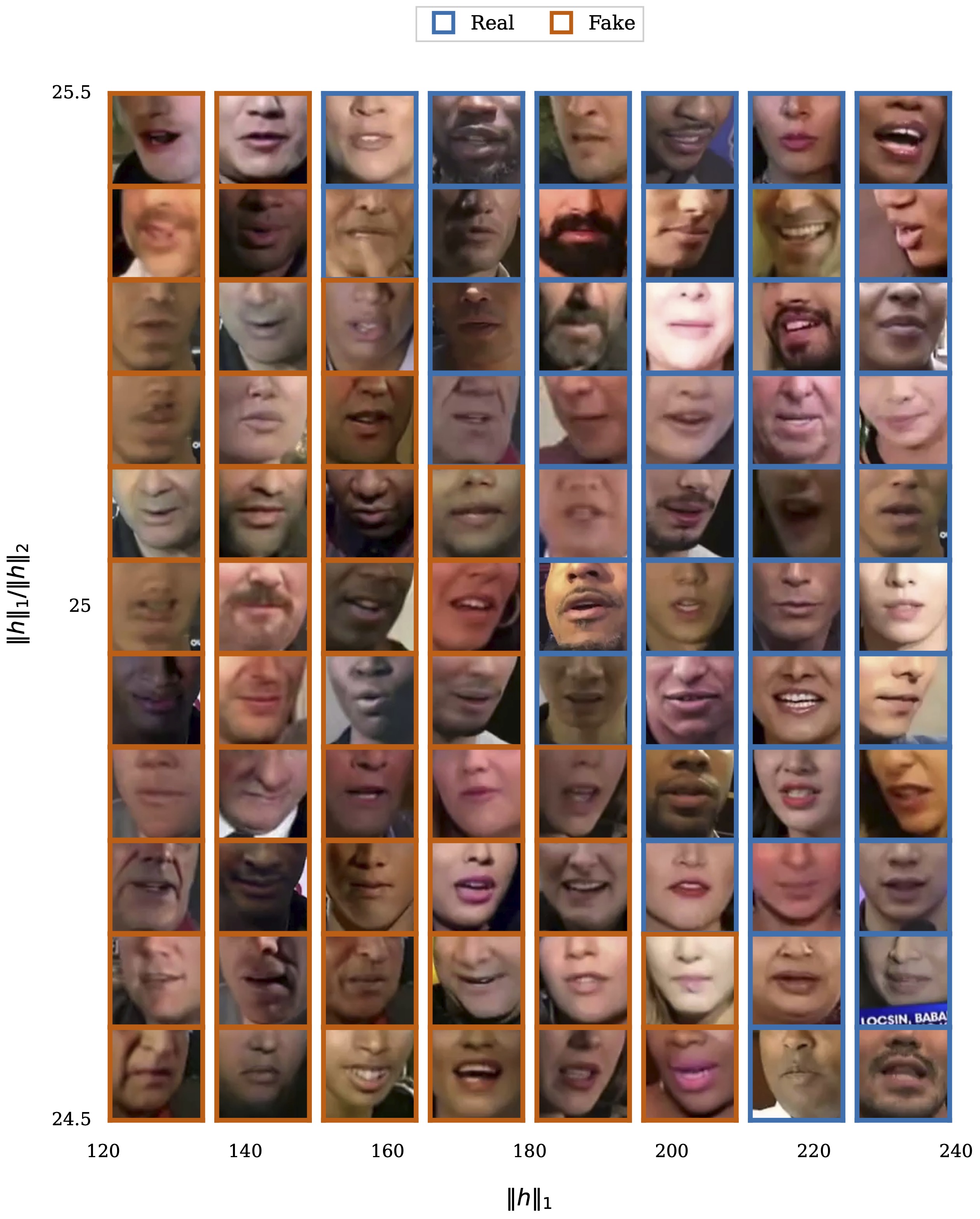}

    \caption{Qualitative examples from FakeAVCeleb, ordered by the NormFake scores. The x-axis corresponds to the magnitude score ($\ell_1$ norm), while the y-axis corresponds to the sparsity score ($\ell_1/\ell_2$ ratio). The scores are computed from AV-HuBERT features. Frames with blue borders denote real samples, while those with orange borders denote fake samples.}
    \label{fig:qualitative_av_hubert}
\end{figure*}

\begin{figure*}[t]
    \centering
    \includegraphics[width=0.85\linewidth]{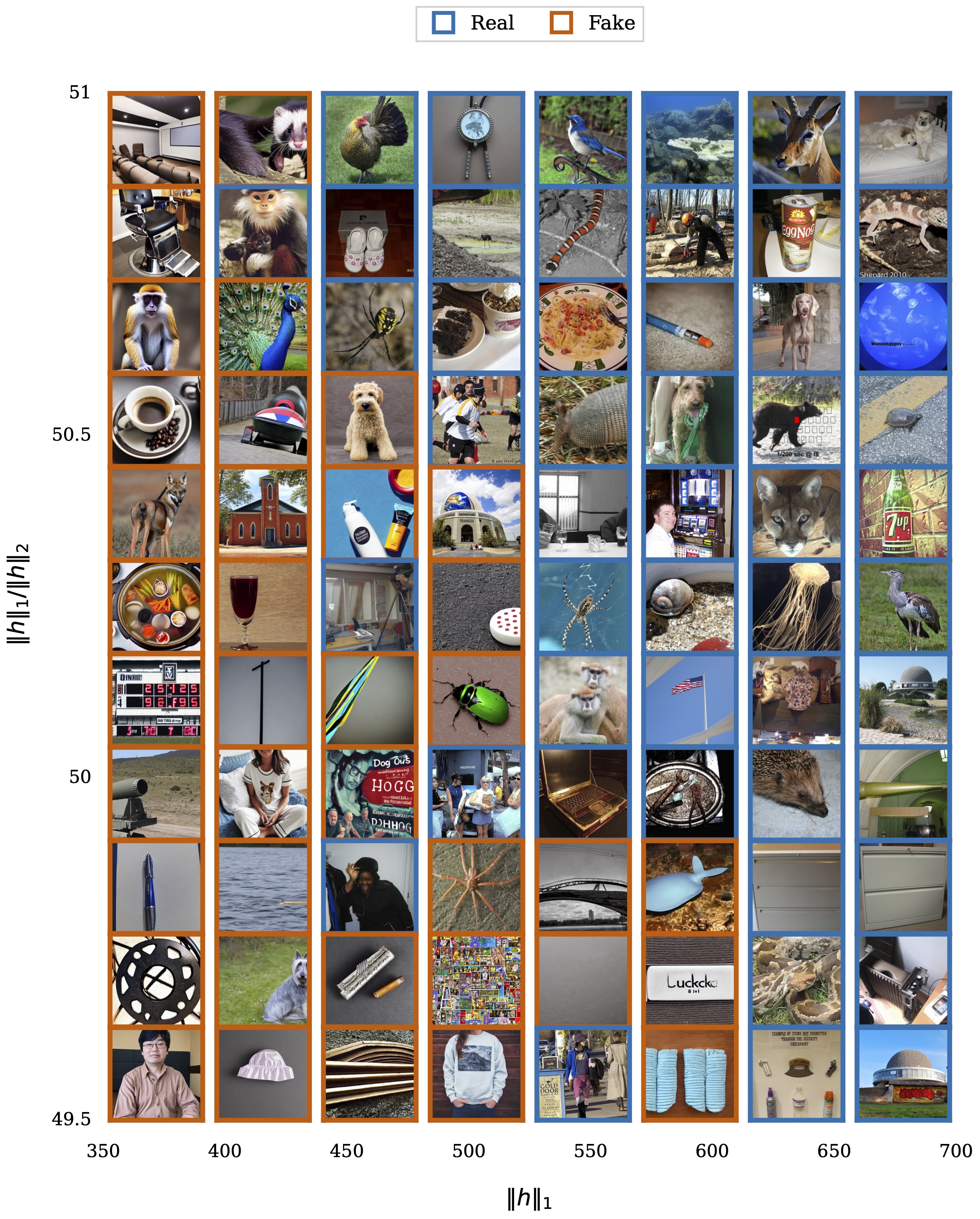}

    \caption{Qualitative examples from GenImage, ordered by the NormFake scores. The x-axis corresponds to the magnitude score ($\ell_1$ norm), while the y-axis corresponds to the sparsity score ($\ell_1/\ell_2$ ratio). The scores are computed from DINOv3-7B features. Images with blue borders denote real samples, while those with orange borders denote fake samples.}
    \label{fig:qualitative_dino}
\end{figure*}

To better understand which image properties influence the magnitude and the sparsity of the representations, we provide qualitative examples in Figure~\ref{fig:qualitative_av_hubert} for AV-HuBERT features on FakeAVCeleb and in Figure~\ref{fig:qualitative_dino} for DINOv3-7B features on GenImage.
The samples are ordered by the magnitude ($\ell_1$ norm) of their representations along the x-axis and by the sparsity (the ratio between $\ell_1$ and $\ell_2$ norms) ratio along the y-axis. 

For AV-HuBERT, we observe a tendency for blurred frames and those containing visible visual artifacts to receive lower scores for both metrics, whereas sharper, more natural-looking frames often receive higher scores.

For DINOv3, images with high-magnitude norms are predominantly semantically coherent photographs containing a single well-defined object or scene, whereas low-magnitude images more frequently correspond to cluttered scenes and unusual viewpoints and framing.
In contrast, the $\ell_1/\ell_2$ ratio exhibits a much weaker visual trend.
Nevertheless, there appears to be a slight tendency for higher norm ratios to correspond to object-centric images, while lower ratios are indicate more complex scenes.

